\documentclass[10pt,journal,compsoc]{IEEEtran}

\ifCLASSOPTIONcompsoc
  \usepackage[nocompress]{cite}
\else
  \usepackage{cite}
\fi
\ifCLASSINFOpdf
   \usepackage[pdftex]{graphicx}
\else
\fi

\usepackage{algorithm}
\usepackage[noend]{algpseudocode}
\usepackage{amsmath}
\usepackage{amssymb}
\usepackage{multirow}
\usepackage{color}
\usepackage{subcaption}

\newcommand{\blue}[1]{\textcolor{black}{#1}}

\begin{document}
\title{Identifying Emotions from Walking Using Affective and Deep Features}
\author{Tanmay~Randhavane,
    Uttaran~Bhattacharya,
    Kyra~Kapsaskis,
    Kurt~Gray,
    Aniket~Bera,
    and~Dinesh~Manocha
\IEEEcompsocitemizethanks{\IEEEcompsocthanksitem T. Randhavane is with the Department of Computer Science, University of North Carolina, Chapel Hill, NC, 27514.\protect\\
E-mail: tanmay@cs.unc.edu
\IEEEcompsocthanksitem K. Kapsaskis and K. Gray are with University of North Carolina at Chapel Hill.
\IEEEcompsocthanksitem A. Bera, U. Bhattacharya, and D. Manocha are with University of Maryland at College Park.}
}


\IEEEtitleabstractindextext{%
\begin{abstract}
We present a new data-driven model and algorithm to identify the perceived emotions of individuals based on their walking styles. Given an RGB video of an individual walking, we extract his/her walking gait in the form of a series of 3D poses. Our goal is to exploit the gait features to classify the emotional state of the human into one of four emotions: happy, sad, angry, or neutral. Our perceived emotion \blue{identification} approach uses deep features learned via LSTM on labeled emotion datasets. Furthermore, we combine these features with affective features computed from gaits using posture and movement cues. These features are classified using a Random Forest Classifier. We show that our mapping between the combined feature space and the perceived emotional state provides $80.07\%$ accuracy in identifying the perceived emotions. In addition to \blue{identifying} discrete categories of emotions, our algorithm also predicts the values of perceived valence and arousal from gaits. We also present an ``\textit{EWalk} (Emotion Walk)" dataset that consists of videos of walking individuals with gaits and labeled emotions. To the best of our knowledge, this is the first gait-based model to identify perceived emotions from videos of walking individuals.
\end{abstract}

}

\maketitle

\IEEEraisesectionheading{\section{Introduction}\label{sec:introduction}}



\IEEEPARstart{E}{motions} play a large role in our lives, defining our experiences and shaping how we view the world and interact with other humans. Perceiving the emotions of social partners helps us understand their behaviors and decide our actions towards them. For example, people communicate very differently with someone they perceive to be angry and hostile than they do with someone they perceive to be calm and content. Furthermore, the emotions of unknown individuals can also govern our behavior, (e.g., emotions of pedestrians at a road-crossing or emotions of passengers in a train station). Because of the importance of perceived emotion in everyday life, automatic emotion recognition is a critical problem in many fields such as games and entertainment, security and law enforcement, shopping, human-computer interaction, human-robot interaction, etc. 

Humans perceive the emotions of other individuals using verbal and non-verbal cues. Robots and AI devices that possess speech understanding and natural language processing capabilities are better at interacting with humans. Deep learning techniques can be used for speech emotion recognition and can facilitate better interactions with humans~\cite{devillers2015inference}.

Understanding the perceived emotions of individuals using non-verbal cues is a challenging problem. \blue{Humans use the non-verbal cues of facial expressions and body movements to perceive emotions.} With a more extensive availability of data, considerable research has focused on using facial expressions to understand emotion~\cite{fabian2016emotionet}. However, recent studies in psychology question the communicative purpose of facial expressions and doubt the quick, automatic process of perceiving emotions from these expressions~\cite{russell2003facialandvocal}. There are situations when facial expressions can be unreliable, such as with ``mock" or ``referential expressions"~\cite{ekman1993facialexpression}. Facial expressions can also be unreliable depending on whether an audience is present~\cite{fernandezdols1995}.

\begin{figure}[t]
    \centering
    \includegraphics[width=0.48\textwidth]{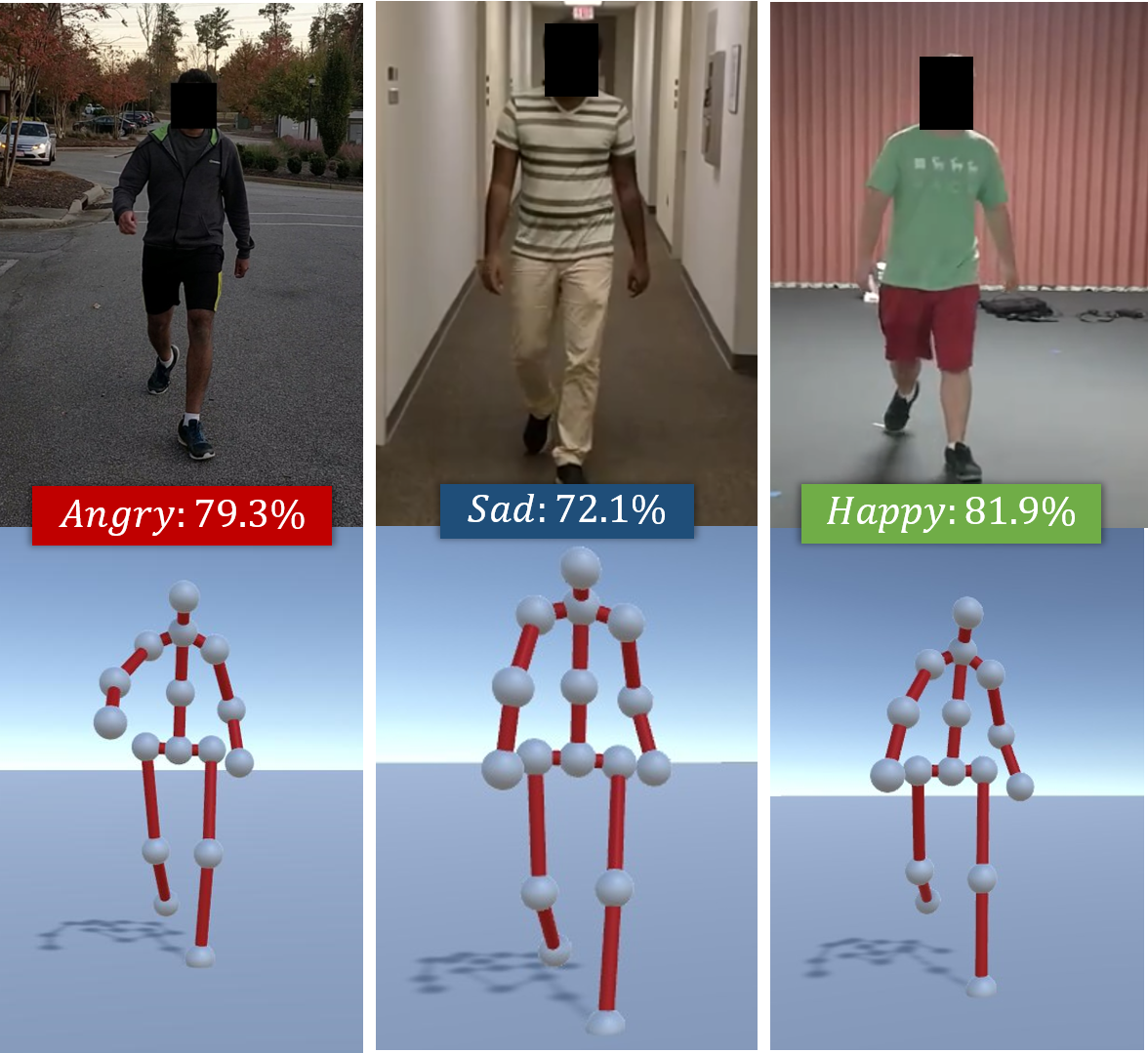}
    \caption{\textbf{Identiying Perceived Emotions}: We present a novel algorithm to identify the perceived emotions of individuals based on their walking styles. Given an RGB video of an individual walking (top), we extract his/her walking gait as a series of 3D poses (bottom). We use a combination of deep features learned via an LSTM and affective features computed using posture and movement cues to then classify into basic emotions (e.g., happy, sad, etc.) using a Random Forest Classifier.}
    \label{fig:cover}
\end{figure}

Research has shown that body expressions are also crucial in emotion expression and perception~\cite{kleinsmith2013affective}. For example, when presented with bodies and faces that expressed either anger or fear (matched correctly with each other or as mismatched compound images), observers are biased towards body expression~\cite{meeren2005rapid}. Aviezer et al.'s study~\cite{aviezer2012} on positive/negative valence in tennis players showed that faces alone were not a diagnostic predictor of valence, but the body alone or the face and body together can be predictive.

Specifically, body expression in walking, or an individual's gait, has been proven to aid in the perception of emotions. In an early study by Montepare et al.~\cite{montepare1987identification}, participants were able to identify sadness, anger, happiness, and pride at a significant rate by observing affective features such as increased arm swinging, long strides, a greater foot landing force, and erect posture. Specific movements have also been correlated with specific emotions. For example, sad movements are characterized by a collapsed upper body and low movement activity~\cite{wallbott1998}. Happy movements have a faster pace with more arm swaying~\cite{michalak2009embodiment}.

\textbf{Main Results:} We present an automatic emotion identification approach for videos of walking individuals (Figure~\ref{fig:cover}). We classify walking individuals from videos into happy, sad, angry, and neutral emotion categories. These emotions represent emotional states that last for an extended period and are more abundant during walking~\cite{ma2006motion}. We extract gaits from walking videos as 3D  poses. We use an LSTM-based approach to obtain deep features by modeling the long-term temporal dependencies in these sequential 3D human poses. We also present spatiotemporal \textit{affective features} representing the posture and movement of walking humans. We combine these affective features with LSTM-based deep features and use a Random Forest Classifier to classify them into four categories of emotion. We observe an improvement of $13.85\%$ in the classification accuracy over other gait-based perceived emotion classification algorithms (Table~\ref{tab:accuracySota}). \blue{We refer to our LSTM-based model between affective and deep features and the perceived emotion labels as our novel data-driven mapping.} 

We also present a new dataset, \textit{``Emotion  Walk  (EWalk),"} which contains videos of individuals walking in both indoor and outdoor locations. Our dataset consists of $1384$ gaits and the perceived emotions labeled using Mechanical Turk.

Some of the novel components of our work include:\\
\noindent 1. A novel data-driven mapping between the affective features extracted from a walking video and the perceived emotions.

\noindent 2. A novel emotion identification algorithm that combines affective features and deep features, obtaining $80.07\%$ accuracy.

\noindent 3. A new public domain dataset, \textit{EWalk}, with walking videos, gaits, and labeled emotions.

The rest of the paper is organized as follows. In Section 2, we review the related work in the fields of emotion modeling, bodily expression of emotion, and automatic recognition of emotion using body expressions. In Section 3, we give an overview of our approach and present the affective features. We provide the details of our LSTM-based approach to identifying perceived emotions from walking videos in Section 4. We compare the performance of our method with state-of-the-art methods in Section 5.  We present the \textit{EWalk} dataset in Section 6.

\section{\blue{Related Work}}
\label{sec:RelatedWork}
\blue{In this section, we give a brief overview of previous works on emotion representation, emotion expression using body posture and movement, and automatic emotion recognition.}

\subsection{\blue{Emotion Representation}}
\begin{figure}[t]
    \centering
    \includegraphics[width=0.48\textwidth]{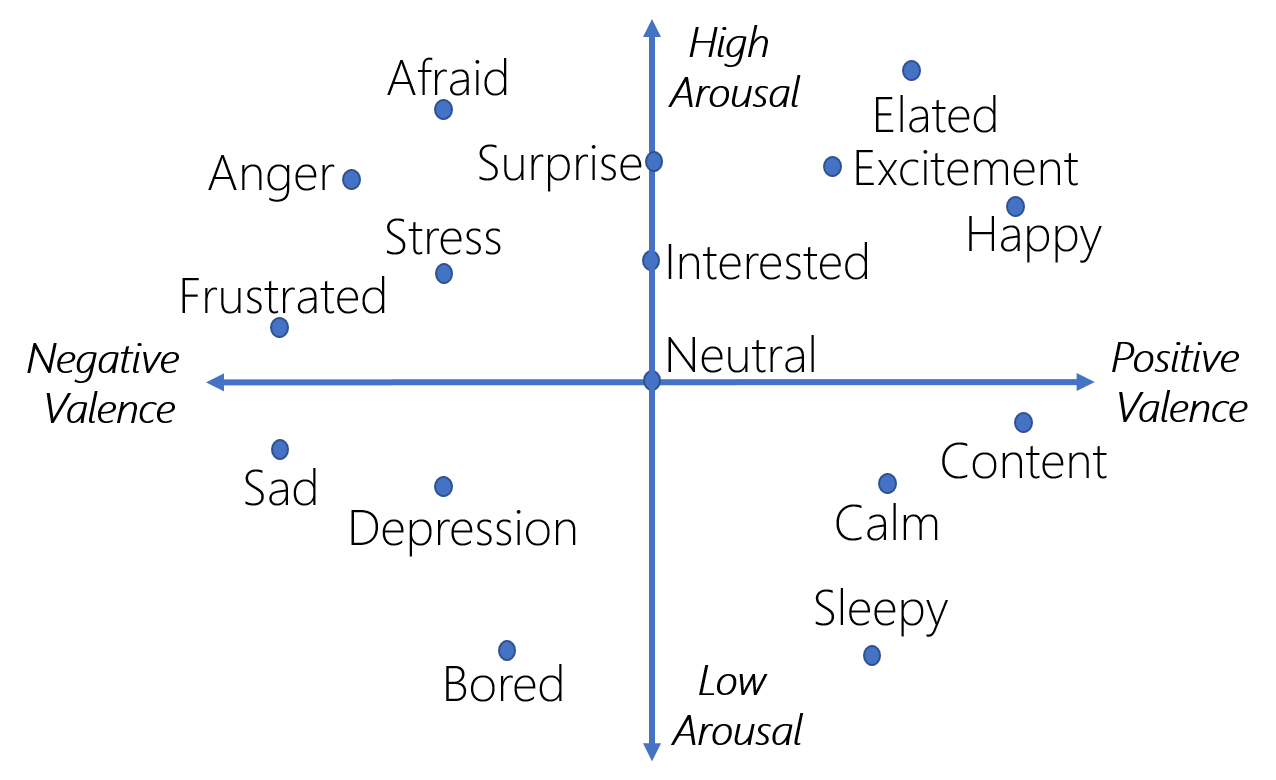}
    \caption{All discrete emotions can be represented by points on a 2D affect space of Valence and Arousal~\protect\cite{loutfi2003social,ekman1967head}.}
  \label{fig:affectSpace}
\end{figure}

\blue{Emotions have been represented using both discrete and continuous representations~\cite{ekman1967head,mehrabian1980basic,kleinsmith2013affective}. In this paper, we focus on discrete representations of the emotions and identify four discrete emotions (happy, angry, sad, and neutral). However, a combination of these emotions can be used to obtain the continuous representation (Section~\ref{sec:affect}). A mapping between continuous representation and the discrete categories developed by Mikels and Morris~\cite{morris1995observations,mikels2005emotional} can be used to predict other emotions.}

\blue{It is important to distinguish between perceived emotions and actual emotions as we discuss the perception of emotions. One of the most obvious cues to another person's emotional state is his or her self-report~\cite{RobinsonClore2002}. However, self-reports are not always available; for example, when people observe others remotely (e.g., via cameras), they do not have the ability to ask about their emotional state. Additionally, self-reports can be imperfect because people can experience an emotion without being aware of it or be unable to translate the emotion into words~\cite{barrett2019emotional}. Therefore, in this paper, we focus on emotions perceived by observers instead of using self-reported measures.}

\blue{Affect expression combines verbal and nonverbal communication, including eye gaze and body expressions, in addition to facial expressions, intonation, and other cues~\cite{picard1998towardagents}.  Facial expressions--like any element of emotional communication--do not exist in isolation. There is no denying that in certain cases, such as with actors and caricatures, it is appropriate to assume affect based on the visual cues from the face, however, in day-to-day life, this does not account for body expressions. More specifically, the way a person walks, or their gait, has been proven to aid in the perception of that person’s emotion~\cite{montepare1987identification}.}

\blue{With the increasing availability of technologies that capture body expression, there is considerable work on the automatic recognition of emotions from body expressions. Most works use a feature-based approach to identify emotion from body expressions automatically. These features are either extracted using purely statistical techniques or using techniques that are inspired by psychological studies. Karg et al.~\cite{karg2013body} surveyed body movement-based methods for automatic recognition and generation of affective expression. Many techniques in this area have focused on activities such as knocking~\cite{gross2010methodology}, dancing~\cite{camurri2003recognizing}, games~\cite{savva2012continuous}, etc. A recent survey discussed various gesture-based emotion recognition techniques~\cite{noroozi2018survey}. These approaches model gesture features (either handcrafted or learned) and then classify these gestures into emotion classes. For example, Piana et al.~\cite{piana2016adaptive,piana2014real,piana2013set,piana2013set} presented methods for emotion recognition from motion-captured data or RGB-D videos obtained from Kinect cameras. Their method is focused on characters that are stationary and are performing various gestures using hands and head joints. They recognize emotions using an SVM classifier that classifies features (both handcrafted and learned) from 3D coordinate and silhouette data. Other approaches have used PCA to model non-verbal movement features for emotion recognition~\cite{de2004modeling,glowinski2011toward}.}

\blue{Laban movement analysis (LMA)~\cite{von1970principles} is a framework for representing the human movement that has widely used for emotion recognition. Many approaches have formulated gesture features based on LMA and used them to recognize emotions~\cite{zacharatos2013emotion,camurri2003recognizing}. The Body Action and Posture Coding System (BAP) is another framework for coding body movement~\cite{dael2012body,huis2014body,van2014body}. Researchers have used BAP to formulate gesture features and used these features for emotion recognition~\cite{dael2012emotion}.}

\blue{Deep learning models have also been used to recognize emotions from non-verbal gesture cues. Sapinski et al.~\cite{sapinski2019emotion} used an LSTM-based network to identify emotions from body movement. Savva et al.~\cite{savva2012continuous} proposed an RNN-based emotion identification method for players of fully-body computer games. Butepage et al.~\cite{butepage2017deep} presented a generative model for human motion prediction and activity classification. Wang et al.~\cite{wang2019learning} proposed an LSTM-based network to recognize pain-related behavior from body expressions. Multimodal approaches that combine cues such as facial expressions, speech, and voice with body expressions have also been proposed~\cite{meeren2005rapid,wagner2011exploring,balomenos2004emotion,caridakis2007multimodal}.}

\blue{In this work, we focus on pedestrians and present an algorithm to recognize emotions from walking using gaits. Our approach uses both handcrafted features (referred to as the affective features), and deep features learned using an LSTM-based network for emotion recognition.}

\subsection{\blue{Automatic Emotion Recognition from Walking}}
\blue{As shown by Montepare et al.~\cite{montepare1987identification}, gaits have the potential to convey emotions. Previous research has shown that gaits can be used to recognize emotions. These approaches have formulated features using gaits obtained as 3D positions of joints from Microsoft Kinect or motion-captured data. For example, Li et al.~\cite{li2016identifying,li2016emotion} used gaits extracted from Microsoft Kinect and recognized the emotions using Fourier Transform and PCA-based features. Roether et al.~\cite{roether2009critical,roether2009features} identified posture and movement features from gaits by conducting a perception experiment. However, their goal was to formulate a set of expressive features and not emotion recognition. Crenn et al.~\cite{crenn2016body} used handcrafted features from gaits and classified them using SVMs. In the following work, they generated neutral movements from expressive movements for emotion recognition. They introduced a cost function that is optimized using the Particle Swarm Optimization method to generate a neutral movement. They used the difference between the expressive and neutral movement for emotion recognition. Karg et al.~\cite{karg2010recognition,karg2009two,karg2009comparison} examined gait information for person-dependent affect recognition using motion capture data of a single walking stride. They formulated handcrafted gait features converted to a lower-dimensional space using PCA and then classified using them SVM, Naive Bayes, fully-connected neural networks. Venture et al.~\cite{venture2014recognizing} used an auto-correlation matrix of the joint angles at each frame of the gait and used similarity indices for classification. Janssen et al.~\cite{janssen2008recognition} used a neural network for emotion identification from gaits and achieved an accuracy of more than $80\%$. However, their method requires special devices to compute 3D ground reaction forces during gait, which may not be available in many situations. Daoudi et al.~\cite{daoudi2017emotion} used a manifold of symmetric positive definite matrices to represent body movement and classified them using the Nearest Neighbors method. Kleinsmith et al.~\cite{kleinsmith2011automatic} used handcrafted features of postures and classified them to recognize affect using a multilayer perceptron automatically. Omlor and Giese~\cite{omlor2007extraction} identified spatiotemporal features that are specific to different emotions in gaits using a novel blind source separation method. Gross et al.~\cite{gross2012effort-shape} performed an Effort-Shape analysis to identify the characteristics associated with positive and negative emotions. They observed features such as walking speed, increased amplitude of joints, thoracic flexion, etc. were correlated with emotions. However, they did not present any method to use these features for automatic identification of emotions.}

\blue{Researchers have also attempted to synthesize gaits with different styles. Ribet et al.~\cite{ribet2019survey} surveyed the literature on style generation and recognition using body movements. Tilmanne et al.~\cite{tilmanne2010expressive} presented methods for generating stylistic gaits using a PCA-based data-driven approach. In the principal component space, they modeled the variability of gaits using Gaussian distributions. In a subsequent work~\cite{tilmanne2012stylistic}, they synthesized stylistic gaits using a method based on Hidden Markov Models (HMMs). This method is based on using neutral walks and modifying them to generate different styles. Troje~\cite{troje2002decomposing} proposed a framework to classify gender and also used it to synthesize new motion patterns. However, the model is not used for emotion identification. Felis et al.~\cite{felis2013modeling} used objective functions to add an emotional aspect to gaits. However, their results only showed sad and angry emotions.}

\blue{Similar to these approaches of emotion recognition from gaits, we use handcrafted features (referred to as affective features). In contrast to previous methods that either use handcrafted features or deep learning methods, we combine the affective features with deep features extracted using an LSTM-based network. We use the resulting joint features for emotion recognition.}

\section{Approach}
\begin{figure*}[t]
  \centering
  \includegraphics[width =\linewidth]{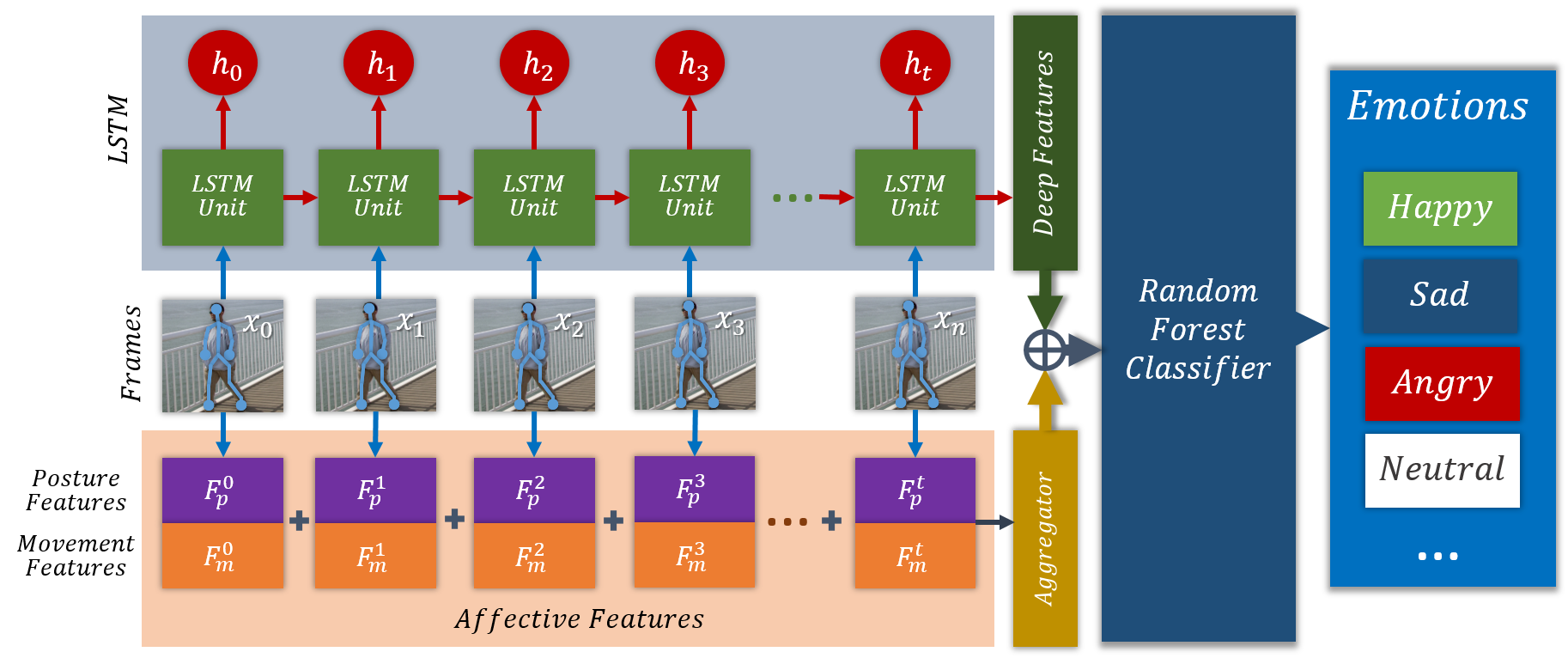}
  \caption{\textbf{Overview:} Given an RGB video of an individual walking, we use a state-of-the-art 3D human pose estimation technique~\protect\cite{dabral2018learning} to extract a set of 3D poses. These 3D poses are passed to an LSTM network to extract deep features. We train this LSTM network using multiple gait datasets. We also compute affective features consisting of both posture and movement features using psychological characterization. We concatenate these affective features with deep features and classify the combined features into $4$ basic emotions using a Random Forest classifier.}
  \label{fig:overview}
\end{figure*}

In this section, we describe our algorithm (Figure~\ref{fig:overview}) for identifying perceived emotions from RGB videos.

\subsection{Notation}
\begin{figure}
    \centering
    \includegraphics[width=0.20\textwidth]{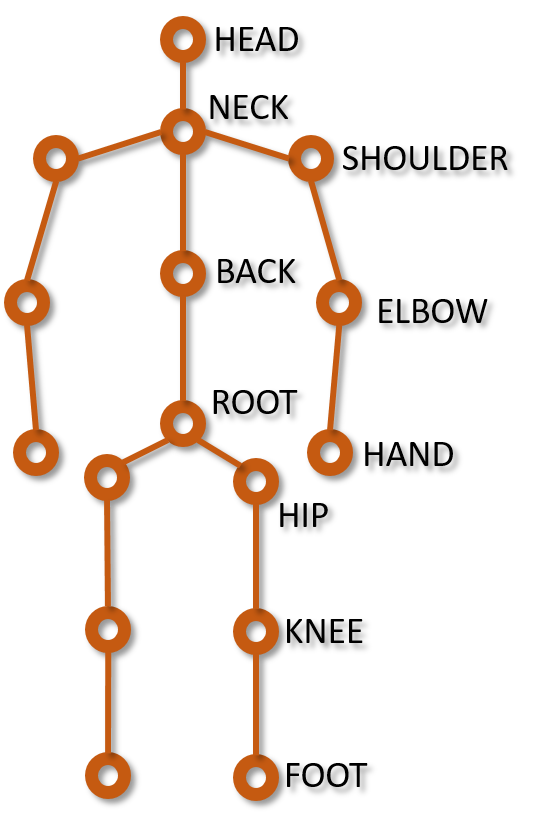}
    \caption{\textbf{Human Representation}: We represent a human by a set of $16$ joints. The overall configuration of the human is defined using these joint positions and is used to extract the features.}
    \label{fig:skeleton}
\end{figure}

For our formulation, we represent a human with a set of $16$ joints, as shown in~\cite{dabral2018learning} (Figure~\ref{fig:skeleton}). A pose $P \in \mathbb{R}^{48}$ of a human is a set of 3D positions of each joint $j_i, i \in \{1,2, ..., 16\}$. For any RGB video $V$, we represent the gait extracted using 3D pose estimation as $G$. The gait $G$ is a set of 3D poses ${P_1, P_2,..., P_{\tau}}$ where $\tau$ is the number of frames in the input video $V$. We represent the extracted affective features of a gait $G$ as $F$. Given the gait features $F$, we represent the predicted emotion by $e \in \{ happy, angry, sad, neutral\}$. \blue{While neutral is not an emotion, it is still a valid state to be used for classification. In the rest of the paper, we refer to these four categories, including neutral as four emotions for convenience.} The four basic emotions represent emotional states that last for an extended period and are more abundant during walking~\cite{ma2006motion}. \blue{A combination of these four emotions can be used to predict affective dimensions of valence and arousal and also other emotions~\cite{mikels2005emotional}.}

\subsection{Overview}
Our real-time perceived emotion prediction algorithm is based on a data-driven approach. We present an overview of our approach in Figure~\ref{fig:overview}. During the offline training phase, we use multiple gait datasets \blue{(described in Section~\ref{sec:mocapDatasets})} and extract affective features. These affective features are based on psychological characterization~\cite{crenn2016body,karg2010recognition} and consist of both posture and movement features. We also extract deep features by training an LSTM network. We combine these deep and affective features and train a Random Forest classifier. At runtime, given an RGB video of an individual walking, we extract his/her gait in the form of a set of 3D poses using a state-of-the-art 3D human pose estimation technique~\cite{dabral2018learning}. We extract affective and deep features from this gait and identify the perceived emotion using the trained Random Forest classifier. We now describe each component of our algorithm in detail.

\subsection{Affective Feature Computation}\label{sec:featureExtraction}
\begin{table}[t]
\caption{\textbf{Posture Features}: We extract posture features from an input gait using emotion characterization in visual perception and psychology literature~\protect \cite{karg2010recognition,crenn2016body}.}
\centering
\begin{tabular}{|l|l|}
\hline
\multicolumn{1}{|c|}{Type}                                                         & \multicolumn{1}{c|}{Description} \\ \hline
Volume                                                                             & Bounding box \\ \cline{1-2}
\multirow{5}{*}{Angle}                                                             & At neck by shoulders \\ \cline{2-2}
                                                                                   & \begin{tabular}[c]{@{}l@{}}At right shoulder by \\ neck and left shoulder\end{tabular} \\ \cline{2-2}
                                                                                   & \begin{tabular}[c]{@{}l@{}}At left shoulder by \\ neck and right shoulder\end{tabular}\\ \cline{2-2}
                                                                                   & At neck by vertical and back  \\ \cline{2-2}
                                                                                   & At neck by head and back  \\ \cline{1-2}
\multirow{5}{*}{Distance}                                                          & \begin{tabular}[c]{@{}l@{}}Between right hand \\ and the root joint\end{tabular} \\ \cline{2-2}
                                                                                   & \begin{tabular}[c]{@{}l@{}}Between left hand \\ and the root joint\end{tabular} \\ \cline{2-2}
                                                                                   & \begin{tabular}[c]{@{}l@{}}Between right foot \\ and the root joint\end{tabular} \\ \cline{2-2}
                                                                                   & \begin{tabular}[c]{@{}l@{}}Between left foot \\ and the root joint\end{tabular} \\ \cline{2-2}
                                                                                   & \begin{tabular}[c]{@{}l@{}}Between consecutive\\  foot strikes (stride length)\end{tabular} \\ \cline{1-2}
\multirow{2}{*}{Area}                                                              & \begin{tabular}[c]{@{}l@{}}Triangle between \\ hands and neck\end{tabular} \\ \cline{2-2}
                                                                                   & \begin{tabular}[c]{@{}l@{}}Triangle between \\ feet and the root joint\end{tabular} \\ \hline
\end{tabular}
\label{tab:posturefeatures}
\end{table}

For an accurate prediction of an individual's affective state, both posture and movement features are essential~\cite{kleinsmith2013affective}. Features in the form of joint angles, distances, and velocities, and space occupied by the body have been used for recognition of emotions and affective states from gaits~\cite{crenn2016body}. Based on these psychological findings, we compute affective features that include both the posture and the movement features. 

We represent the extracted affective features of a gait $\textbf{G}$ as a vector $F \in \mathbb{R}^{29}$. 
For feature extraction, we use a single stride from each gait corresponding to consecutive foot strikes of the same foot. We used a single cycle in our experiments because in some of the datasets (CMU, ICT, EWalk) only a single walk cycle was available. When multiple walk cycles are available, they can be used to increase accuracy.

\subsubsection{Posture Features}
We compute the features $F_{p, t} \in \mathbb{R}^{12}$ related to the posture $P_t$ of the human at each frame $t$ using the skeletal representation (computed using TimePoseNet Section~\ref{sec:timeposenet}). We list the posture features in Table~\ref{tab:posturefeatures}. \blue{These posture features are based on prior work by Crenn et al.~\cite{crenn2016body}. They used upper body features such as the area of the triangle between hands and neck, distances between hand, shoulder, and hip joints, and angles at neck and back. However, their formulation does not consider features related to foot joints, which can convey emotions in walking~\cite{roether2009features}. Therefore, we also include areas, distances, and angles of the feet joints in our posture feature formulation.}

We define posture features of the following types:
\begin{itemize}
    \item Volume: According to Crenn et al.~\cite{crenn2016body}, body expansion conveys positive emotions while a person has a more compact posture during negative expressions. We model this by the volume $F_{volume, t} \in \mathbb{R}$ occupied by the bounding box around the human. 
    \item Area: We also model body expansion by areas of triangles between the hands and the neck and between the feet and the root joint $F_{area, t} \in \mathbb{R}^2$.
    \item Distance: Distances between the feet and the hands can also be used to model body expansion $F_{distance, t} \in \mathbb{R}^4$.
    \item Angle: Head tilt is used to distinguish between happy and sad emotions~\cite{crenn2016body,karg2010recognition}. We model this by the angles extended by different joints at the neck $F_{angle, t} \in \mathbb{R}^5$.
\end{itemize}

We also include stride length as a posture feature. Longer stride lengths convey anger and happiness and shorter stride lengths convey sadness and neutrality~\cite{karg2010recognition}. Suppose we represent the positions of the left foot joint $j_{lFoot}$ and the right foot joint $j_{rFoot}$ in frame $t$ as $\vec{p}(j_{lFoot}, t)$ and $\vec{p}(j_{rFoot}, t)$ respectively. Then the stride length $s \in \mathbb{R}$ is computed as:
\begin{eqnarray}
    s = \max\limits_{t \in 1..\tau}||\vec{p}(j_{lFoot}, t) - \vec{p}(j_{rFoot}, t)||
\end{eqnarray}

We define the posture features $F_{p}\in \mathbb{R}^{13}$ as the average of $F_{p, t}, t=\{1,2,..,\tau\}$ combined with the stride length:
\begin{eqnarray}
    F_{p} = \frac{\sum_{t} F_{p, t}}{\tau} \cup s,
\end{eqnarray}

\subsubsection{Movement Features}
\begin{table}[t]
\caption{\textbf{Movement Features}: We extract movement features from an input gait using emotion characterization in visual perception and psychology literature~\protect \cite{karg2010recognition,crenn2016body}.}
\centering
\begin{tabular}{|l|l|c|}
\hline
\multicolumn{1}{|c|}{Type}                                                         & \multicolumn{1}{c|}{Description} \\ \hline
\multirow{5}{*}{Speed}                                                             & Right hand \\ \cline{2-2}
                                                                                   & Left hand  \\ \cline{2-2}
                                                                                   & Head \\ \cline{2-2}
                                                                                   & Right foot \\ \cline{2-2}
                                                                                   & Left foot  \\ \cline{1-2}
\multirow{5}{*}{Acceleration Magnitude} & Right hand  \\ \cline{2-2}
                                                                                   & Left hand \\ \cline{2-2}
                                                                                   & Head \\ \cline{2-2}
                                                                                   & Right foot  \\ \cline{2-2}
                                                                                   & Left foot \\ \cline{1-2}
\multirow{5}{*}{Movement Jerk}        & Right hand \\ \cline{2-2}
                                                                                   & Left hand  \\ \cline{2-2}
                                                                                   & Head  \\ \cline{2-2}
                                                                                   & Right foot  \\ \cline{2-2}
                                                                                   & Left foot  \\ \cline{1-2}
Time                                                                               & One gait cycle   \\ \hline
\end{tabular}
\label{tab:movementfeatures}
\end{table}

Psychologists have shown that motion is an important characteristic for the perception of different emotions~\cite{gross2012effort-shape}. High arousal emotions are more associated with rapid and increased movements than low arousal emotions. We compute the movement features $F_{m, t} \in \mathbb{R}^{15}$ at frame $t$ by considering the magnitude of the velocity, acceleration, and movement jerk of the hand, foot, and head joints using the skeletal representation. For each of these five joints $j_i, i={1,...,5}$, we compute the magnitude of the first, second, and third finite derivatives of the position vector $\vec{p}(j_i, t)$ at frame $t$. We list the movement features in Table~\ref{tab:movementfeatures}. \blue{These movement features are based on prior work by Crenn et al.~\cite{crenn2016body}. Similar to the posture features, we combine the upper body features from Crenn et al.~\cite{crenn2016body} with lower body features related to feet joints.}

Since faster gaits are perceived as happy or angry whereas slower gaits are considered sad~\cite{karg2010recognition}, we also include the time taken for one walk cycle ($gt\in \mathbb{R}$) as a movement feature. We define the movement features $F_{m}\in \mathbb{R}^{16}$ as the average of $F_{m, t}, t=\{1,2,..,\tau\}$:
\begin{eqnarray}
    F_{m} = \frac{\sum_{t} F_{m, t}}{\tau} \cup gt,
\end{eqnarray}

We combine posture and movement features and define \textbf{affective features} $F$ as: $F = F_{m} \cup  F_{p}$.

\section{Perceived Emotion Identification}
We use a \textit{vanilla} LSTM network~\cite{greff2017lstm} with a cross-entropy loss that models the temporal dependencies in the gait data. We chose an LSTM network to model deep features of walking because it captures the geometric consistency and temporal dependency among video frames for gait modeling~\cite{luo2018lstm}. We describe the details of the training of the LSTM in this section.

\subsection{Datasets}\label{sec:mocapDatasets}
We used the following publicly available datasets for training our perceived emotion classifier: 
\begin{itemize}
    \item \textbf{Human3.6M}~\cite{h36m_pami}: This dataset consists of $3.6$ million 3D human images and corresponding poses. It also contains video recordings of $5$ female and $6$ male professional actors performing actions in $17$ scenarios including taking photos, talking on the phone, participating in discussions, etc. The videos were captured at 50 Hz with four calibrated cameras working simultaneously. We used motion-captured gaits from $14$ videos of the subjects walking from this dataset.
    \item \textbf{CMU}~\cite{CMUGait}: The CMU Graphics Lab Motion Capture Database contains motion-captured videos of humans interacting among themselves (\textit{e.g.}, talking, playing together), interacting with the environment (\textit{e.g.}, playgrounds, uneven terrains), performing physical activities (\textit{e.g.}, playing sports, dancing), enacting scenarios (\textit{e.g.}, specific behaviors), and locomoting (\textit{e.g.}, running, walking). In total, there are motion captured gaits from $49$ videos of subjects walking with different styles.
    \item \textbf{ICT}~\cite{narang2017motion}: This dataset contains motion-captured gaits from walking videos of $24$ subjects. The videos were annotated by the subjects themselves, who were asked to label their own motions as well as motions of other subjects familiar to them.
    \item \textbf{BML}~\cite{ma2006motion}: This dataset contains motion-captured gaits from $30$ subjects (15 male and 15 female). The subjects were nonprofessional actors, ranging between $17$ and $29$ years of age with a mean age of $22$ years. For the walking videos, the actors walked in a triangle for 30 sec, turning clockwise and then counterclockwise in two individual conditions. Each subject provided $4$ different walking styles in two directions, resulting in $240$ different gaits.
    \item \textbf{SIG}~\cite{xia2015realtime}: This is a dataset of $41$ synthetic gaits generated using local mixtures of autoregressive (MAR) models to capture the complex relationships between the different styles of motion. The local MAR models were developed in real-time by obtaining the nearest examples of given pose inputs in the database. The trained model were able to adapt to the input poses with simple linear transformations. Moreover, the local MAR models were able to predict the timings of synthesized poses in the output style.
    \item \textbf{EWalk (Our novel dataset)}: We also collected videos and extracted $1136$ gaits using 3D pose estimation. We present details about this dataset in Section~\ref{sec:dataset}.
\end{itemize}

\begin{figure}
  \centering
  \includegraphics[width=0.48\textwidth]{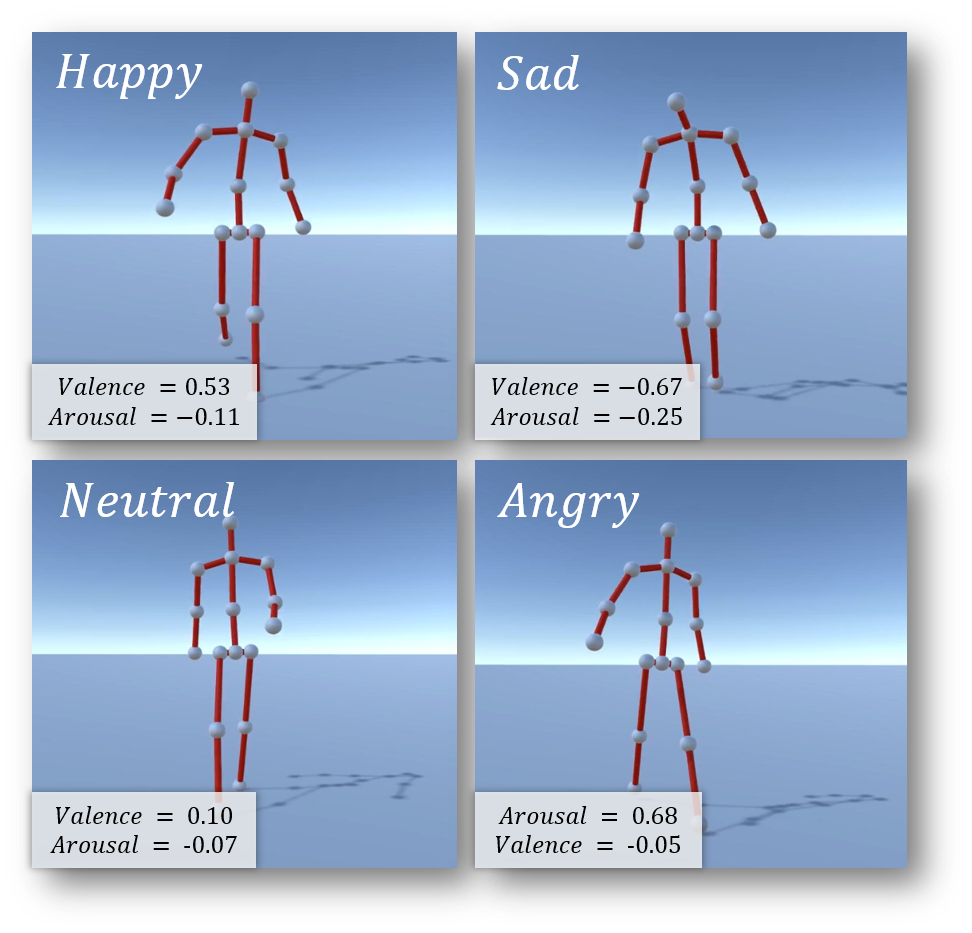}
  \caption{\textbf{Gait Visualizations:} We show the visualization of the motion-captured gaits of four individuals with their classified emotion labels. Gait videos from $248$ motion-captured gaits were displayed to the participants in a web based user study to generate labels. We use that data for training and validation.}
  \label{fig:gaitVideo}
\end{figure}

The wide variety of these datasets includes acted as well as non-acting and natural-walking datasets (CMU, ICT) where the subjects were not told to assume an emotion. These datasets provide a good sample of real-world scenarios. \blue{For the acted video datasets, we did not use the acted emotion labels for the gaits, but instead obtained the perceived emotion labels with a user study (Section~\ref{sec:labeling}).}

\subsection{Perceived Emotion Labeling}\label{sec:labeling}
We obtained the perceived emotion labels for each gait using a web-based user study. 

\subsubsection{Procedure}
We generated visualizations of each motion-captured gait using a skeleton mesh (Figure~\ref{fig:gaitVideo}). For the \textit{EWalk} dataset, we presented the original videos to the participants when they were available. We hid the faces of the actors in these videos to ensure that the emotions were perceived from the movements of the body and gaits, not from the facial expressions. 

\subsubsection{Participants}
We recruited $688$ participants ($279$ female, $406$ male, $\overline{age} = 34.8$) from Amazon Mechanical Turk and the participant responses  were used to generate perceived emotion labels. Each participant watched and rated $10$ videos from one of the datasets. The videos were presented randomly and for each video we obtained a minimum of $10$ participant responses. \blue{Participants who provided a constant rating to all gaits or responded in a very short time were ignored for further analysis ($3$ participants total).}


\subsubsection{Analysis}
We asked each participant whether he/she perceived the gait video as happy, angry, sad, or neutral on 5-point Likert items ranging from Strongly Disagree to Strongly Agree. For each gait $\textbf{G}_i$ in the datasets, we calculated the mean of all participant responses ($r^{e}_{i, j}$) to each emotion:
\begin{eqnarray}
    r^{e}_i = \frac{\sum_{j=1}^{n_p} r^{e}_{i,j}}{n_p},
\end{eqnarray}
where $n_p$ is the number of participant responses collected and $e$ is one of the four emotions: angry, sad, happy, neutral. 

We analyzed the correlation between participants' responses to the questions relating to the four emotions (Table~\ref{tab:correl}). A correlation value closer to $1$ indicates that the two variables are positively correlated and a correlation value closer to $-1$ indicates that the two variables are negatively correlated. A correlation value closer to $0$ indicates that two variables are uncorrelated. As expected, \textit{happy} and \textit{sad} are negatively correlated and \textit{neutral} is uncorrelated with the other emotions.

\begin{table}[t]
\centering
\caption{\textbf{Correlation Between Emotion Responses}: We present the correlation between participants' responses to questions relating to the four emotions.}
\begin{tabular}{|c|c|c|c|c|}
\hline
        & Happy  & Angry  & Sad    & Neutral \\ \hline
Happy   & 1.000  & -0.268 & -0.775 & -0.175  \\ \hline
Angry   & -0.268 & 1.000  & -0.086 & -0.058  \\ \hline
Sad     & -0.775 & -0.086 & 1.000  & -0.036  \\ \hline
Neutral & -0.175 & -0.058 & -0.036 & 1.000   \\ \hline
\end{tabular}
\label{tab:correl} 
\end{table}

Previous research in the psychology literature suggests that social perception is affected by the gender of the observer~\cite{carli1995nonverbal,forlizzi2007interface,kramer2016closing}. To verify that our results do not significantly depend on the gender of the participants, we performed a t-test for differences between the responses by male and female participants. We observed that the gender of the participant did not affect the responses significantly ($t = -0.952, p = 0.353$).

We obtained the emotion label $e_i$ for $\textbf{G}_i$ as follows:
\begin{eqnarray}
    e_i = e \mid r^{e}_i > \theta,
\end{eqnarray}
where $\theta = 3.5$ is an experimentally determined threshold for emotion perception.

If there are multiple emotions with average participant responses greater than $ r^{e}_i > \theta$, the gait is not used for training.

\subsection{Long Short-Term Memory (LSTM) Networks}
LSTM networks~\cite{greff2017lstm} are neural networks with special units known as ``memory cells'' that can store data values from particular time steps in a data sequence for arbitrarily long time steps. Thus, LSTMs are useful for capturing temporal patterns in data sequences and subsequently using those patterns in prediction and classification tasks. To perform supervised classification, LSTMs, like other neural networks, are trained with a set of training data and corresponding class labels. However, unlike traditional feedforward neural networks that learn structural patterns in the training data, LSTMs learn feature vectors that encode temporal patterns in the training data.

LSTMs achieve this by training one or more ``hidden'' cells, where the output at every time step at every cell depends on the current input and the outputs at previous time steps. These inputs and outputs to the LSTM cells are controlled by a set of gates. LSTMs commonly have three kinds of gates: the input gate, the output gate, and the forget gate, represented by the following equations:

\begin{align}
    \textit{Input Gate $(i)$:} \qquad & i_t = \sigma(W_i + U_ih_{t-1} + b_i) \\
    \textit{Output Gate $(o)$:} \qquad & o_t = \sigma(W_o + U_oh_{t-1} + b_o) \\
    \textit{Forget Gate $(f)$:} \qquad & f_t = \sigma(W_f + U_fh_{t-1} + b_f)
\end{align}
where $\sigma(\cdot)$ denotes the activation function and $W_g$, $U_g$ and $b_g$ denote the weight \blue{matrix} for the input at the current time step, the weight matrix for the hidden cell at the previous time step, and the bias, on gate $g\in\{i, o, f\}$, respectively. Based on these gates, the hidden cells in the LSTMs are then updated using the following equations:

\begin{align}
    c_t &= f_t \circ c_{t-1} + i_t \circ \sigma(W_cx_t + U_ch_t + b_c) \\
    h_t &= \sigma(o_t \circ c_t)
\end{align}
where $\circ$ denotes the Hadamard or elementwise product, $c$ is referred to as the cell state, and $W_c$, $U_c$ and $b_c$ are the weight matrix for the input at the current time step, the weight matrix for the hidden cell at the previous time step, and the bias, on $c$, respectively.

\subsection{Deep Feature Computation}
We used the LSTM network shown in Figure~\ref{fig:overview}. We obtained deep features from the final layer of the trained LSTM network. We used the $1384$ gaits from the various public datasets (Section~\ref{sec:mocapDatasets}). We also analyzed the extracted deep features using an LSTM encoder-decoder architecture with reconstruction loss. We generated synthetic gaits and observed that our LSTM-based deep features correctly model the 3D positions of joints relative to each other at each frame. The deep features also capture the periodic motion of the hands and legs. 

\subsubsection{Implementation Details}
The training procedure of the LSTM network that we followed is laid out in Algorithm~\ref{algo:LSTM_net}. For training, we used a mini-batch size of $8$ (\textit{i.e.}, $b=8$ in Algorithm~\ref{algo:LSTM_net}) and $500$ training epochs. We used the Adam optimizer~\cite{adam} with an initial learning rate of $0.1$, decreasing it to $\frac{1}{10}$-th of its current value after $250$, $375$, and $438$ epochs. We also used a momentum of $0.9$ and a weight-decay of $5\times 10^{-4}$. The training was carried out on an NVIDIA GeForce GTX 1080 Ti GPU.

\begin{algorithm}
\caption{LSTM Network for Emotion Perception}\label{algo:LSTM_net}
\hspace*{\algorithmicindent} \textbf{Input:} $N$ training gaits $\{\textbf{G}_i\}_{i=1\dots N}$ and corresponding emotion labels $\{\text{L}_i\}_{i=1\dots N}$.\\
\hspace*{\algorithmicindent} \textbf{Output:} Network parameters $\theta$ such that the loss $\sum_{i=1}^N\lVert\text{L}_i - f_{\theta}(\textbf{G}_i)\rVert^2$ is minimized, where $f_\theta(\cdot)$ denotes the network.
\begin{algorithmic}[1]
\Procedure{Train}{}
\For {number of training epochs}
\For {number of iterations per epoch}
\State Sample mini-batch of $b$ training gaits and corresponding labels
\State Update the network parameters $\theta$ w.r.t. the $b$ samples using backpropagation.
\EndFor
\EndFor
\EndProcedure
\end{algorithmic}
\end{algorithm}

\subsection{Classification}\label{sec:classifier}
We concatenate the deep features with affective features and use a Random Forest classifier to classify these concatenated features. Before combining the affective features with the deep features, we normalize them to a range of $[-1, 1]$. We use Random Forest Classifier with $10$ estimators and a maximum depth of $5$. We use this trained classifier to classify perceived emotions. \blue{We refer to this trained classifier as our novel data-driven mapping between affective features and perceived emotion.}

\subsection{Realtime Perceived Emotion Recognition}\label{sec:timeposenet}
At runtime, we take an RGB video as input and use the trained classifier to identify the perceived emotions. \blue{We exploit a real-time 3D human pose estimation algorithm, \textit{TimePoseNet}~\cite{dabral2018learning} to extract 3D joint positions from RGB videos. \textit{TimePoseNet} uses a semi-supervised learning method that utilizes the more widely available 2D human pose data~\cite{lin2014microsoft} to learn the 3D information.}

\textit{TimePoseNet} is a single person model and expects a sequence of images cropped closely around the person as input. Therefore, we first run a real-time person detector~\cite{cao2017realtime} on each frame of the RGB video and extract a sequence of images cropped closely around the person in the video $V$. The frames of the input video $V$ are sequentially passed to \textit{TimePoseNet}, which computes a 3D pose output for each input frame. The resultant poses ${P_1, P_2,..., P_{\tau}}$ represent the extracted output gait $G$. We normalize the output poses so that the root position always coincides with the origin of the 3D space. We extract features of the gait $G$ using the trained LSTM model. We also compute the affective features and classify the combined features using the trained Random Forest classifier.
\section{Results}
We provide the classification results of our algorithm in this section.

\subsection{Analysis of Different Classification Methods}
\blue{We compare different classifiers to classify our combined deep and affective features and compare the resulting accuracy values in Table~\ref{tab:accuracyMethods}. These results are computed using 10-fold cross-validation on $1384$ gaits in the gait datasets described in Section~\ref{sec:mocapDatasets}. We compared Support Vector Machines (SVMs) with both linear and RBF kernel and Random Forest methods. The SVMs were implemented using the one-vs-one approach for multi-class classification. The Random Forest Classifier was implemented with $10$ estimators and a maximum depth of $5$.} We use the Random Forest classifier in the subsequent results because it provides the highest accuracy ($80.07\%$) of all the classification methods. Additionally, our algorithm achieves $79.72\%$ accuracy on the non-acted datasets (CMU and ICT), indicating that it performs equally well on acted and non-acted data.

\begin{table}[t]
    \caption{\textbf{Performance of Different Classification Methods}: We analyze different classification algorithms to classify the concatenated deep and affective features. We observe an accuracy of $80.07\%$ with the Random Forest classifier.}
    \label{tab:accuracyMethods}
    \centering
    \begin{tabular}{|l|l|}
    \hline
    \multicolumn{1}{|c|}{\textbf{Algorithm (Deep + Affective Features)}} & \multicolumn{1}{c|}{\textbf{Accuracy}} \\ \hline
    LSTM + Support Vector Machines (SVM RBF)         & 70.04\%    \\ \hline
    LSTM + Support Vector Machines (SVM Linear)            & 71.01\%    \\ \hline
    \textit{LSTM + Random Forest}  & \textbf{80.07}\%     \\ \hline
    \end{tabular}
\end{table} 

\subsection{Comparison with Other Methods}
In this section, we present the results of our algorithm and compare it with other state-of-the-art methods:
\begin{itemize}
    \item Karg et al.~\cite{karg2010recognition}:  This method is based on using gait features related to shoulder, neck, and thorax angles, stride length, and velocity. These features are classified using PCA-based methods. This method only models the posture features for the joints and doesn't model the movement features.
    \item Venture et al.~\cite{venture2014recognizing}: This method uses the auto-correlation matrix of the joint angles at each frame and uses similarity indices for classification. The method provides good intra-subject accuracy but performs poorly for the inter-subject databases.
    \item Crenn et al.~\cite{crenn2016body}: This method uses affective features from both posture and movement and classifies these features using SVMs. This method is trained for more general activities like knocking and does not use information about feet joints.
    \item Daoudi et al.~\cite{daoudi2017emotion}: This method uses a manifold of symmetric positive definite matrices to represent body movement and classifies them using the Nearest Neighbors method.
    \item Crenn et al.~\cite{crenn2017toward}: This method synthesizes a neutral motion from an input motion and uses the difference between the input and the neutral emotion as the feature for classifying emotions. This method does not use the psychological features associated with walking styles.
    \item Li et al.~\cite{li2016identifying}: This method uses a Kinect to capture the gaits and identifies whether an individual is angry, happy, or neutral using four walk cycles using a feature-based approach. \blue{These features are obtained using Fourier Transform and Principal Component Analysis.}
\end{itemize}

We also compare our results to a baseline where we use the LSTM to classify the gait features into the four emotion classes. Table~\ref{tab:accuracySota} provides the accuracy results of our algorithm and shows comparisons with other methods. These methods require input in the form of 3D human poses and  then they identify the emotions perceived from those gaits. For this experiment, we extracted gaits from the RGB videos of the \textit{EWalk} dataset and then provided them as input to the state-of-the-art methods along with the motion-captured gait datasets. \blue{Accuracy results are obtained using 10-fold cross-validation on $1384$ gaits from the various datasets (Section~\ref{sec:mocapDatasets}). For this evaluation, the gaits were randomly distributed into training and testing sets, and the accuracy values were obtained by averaging over $1000$ random partitions.}

\begin{table}[t]
    \caption{\textbf{Accuracy}: Our method with combined deep and affective features classified with a Random Forest classifier achieves an accuracy of $80.07\%$. We observe an improvement of $13.85\%$ over state-of-the-art emotion identification methods and an improvement of $24.60\%$ over a baseline LSTM-based classifier. \blue{All methods were compared on $1384$ gaits obtained from the datasets described in Section~\ref{sec:mocapDatasets}}}
    \label{tab:accuracySota}
    \centering
    \begin{tabular}{|l|l|}
    \hline
    \multicolumn{1}{|c|}{\textbf{Method}}         & \textbf{Accuracy} \\ \hline
    Baseline (Vanilla LSTM)      & 55.47\%      \\ \hline
    Affective Features Only     & 68.11\%      \\ \hline
    Karg et al.~\cite{karg2010recognition}    & 39.58\%         \\ \hline
    Venture et al.~\cite{venture2014recognizing} &    30.83\%      \\ \hline
    Crenn et al.~\cite{crenn2016body}   &   66.22\%       \\ \hline
    Crenn et al.~\cite{crenn2017toward}   &  40.63\%       \\ \hline
    Daoudi et al.~\cite{daoudi2017emotion}  &   42.52\%       \\ \hline
    Li et al.~\cite{li2016identifying}  &  53.73\%       \\ \hline
    \textit{Our Method (Deep + Affective Features)}    &   \textbf{80.07\%}       \\ \hline
    \end{tabular}
\end{table}

\begin{figure}[t]
    \centering
    \includegraphics[width=\columnwidth]{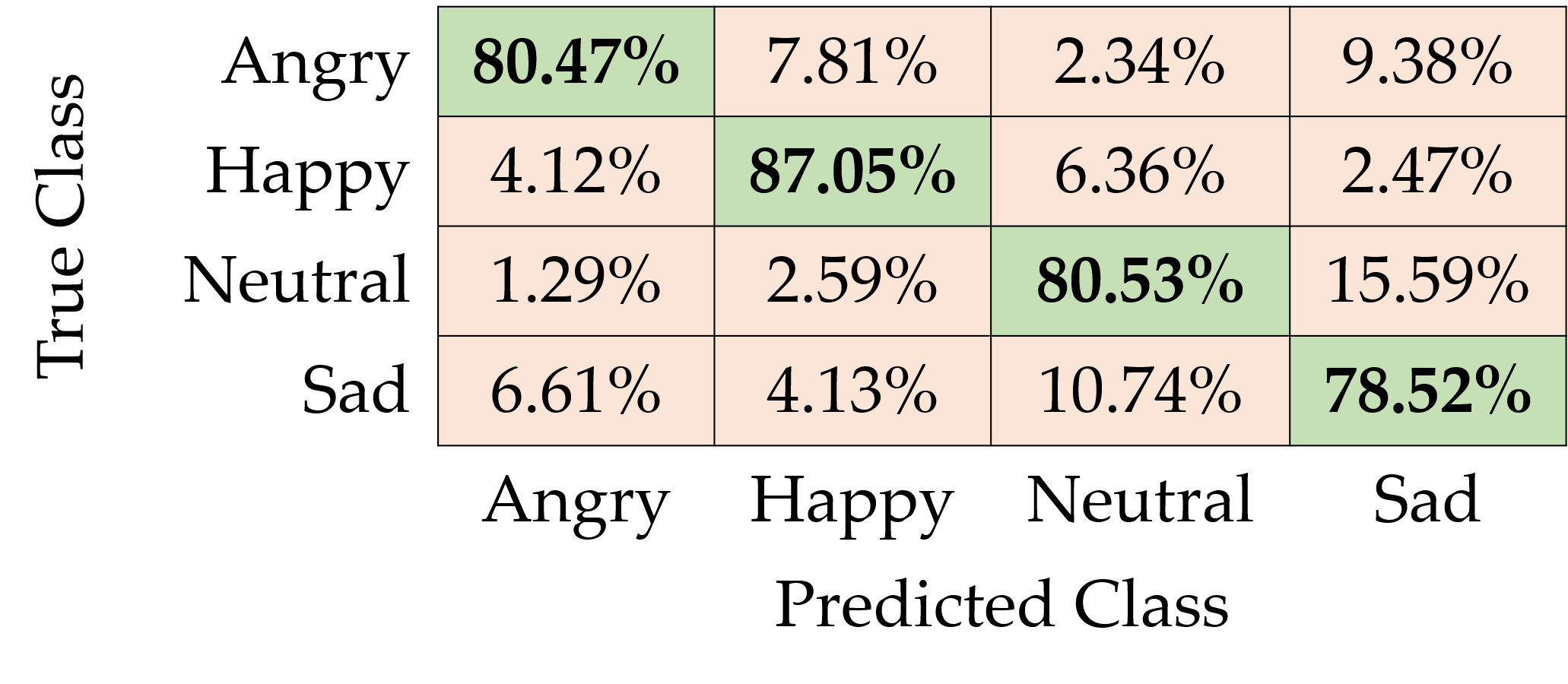}
    \caption{\textbf{Confusion Matrix}: For each emotion class, we show the percentage of gaits belonging to that class that were correctly classified by the LSTM+Random Forest classifier (green background) and the percentage of gaits that were misclassified into other classes (red background).}
    \label{fig:cm_lstm}
\end{figure}

We also show the percentage of gaits that the LSTM+Random Forsest classifier correctly classified for each emotion class in Figure~\ref{fig:cm_lstm}. As we can see, for every class, around $80\%$ of the gaits are correctly classified, implying that the classifier learns to recognize each class equally well. Further, when the classifier does make mistakes, it tends to confuse neutral and sad gaits more than between any other class pairs.

\subsection{Analysis of the Learned Deep Features}

For instance, in section 5.3, "the top 3 principal component directions" is stated, however, this is the first mention of PCA without any details about how, when or where PCA was applied.

\begin{figure}[b]
    \centering
    \includegraphics[width=\columnwidth]{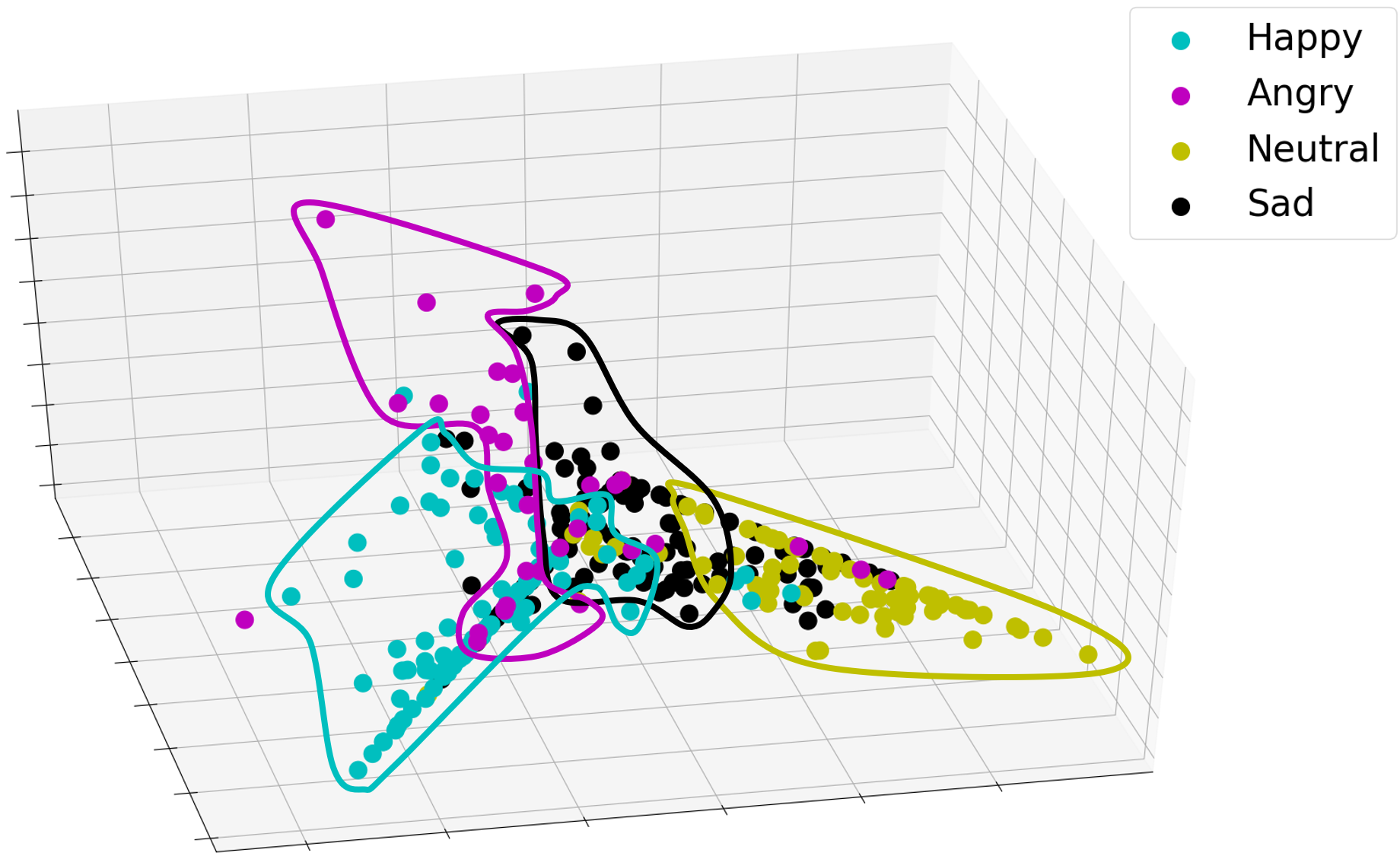}
    \caption{\textbf{Scatter Plot of the Learned Deep Features}: These are the deep features learned by the LSTM network from the input data points, projected in the 3 principal component directions. The different colors correspond to the different input class labels. We can see that the features for the different classes are well-separated in the 3 dimensions. This implies that the LSTM network learns meaningful representations of the input data for accurate classification.}
    \label{fig:scatter_plot}
\end{figure}

We visualize the scatter of the deep feature vectors learned by the LSTM network. \blue{To visualize the $32$ dimensional deep features, we convert them to a 3D space. We use Principal Component Analysis (PCA) and project the features in the top three principal component directions.} This is shown in Figure~\ref{fig:scatter_plot}. We observe that the data points are well-separated even in the projected dimension. By extension, this implies that the deep features are at least as well separated in their original dimension. Therefore, we can conclude that the LSTM network has learned meaningful representations of the input data that help it distinguish accurately between the different emotion classes.

\begin{figure*}[!ht]
    \centering
    \includegraphics[width=0.98\textwidth]{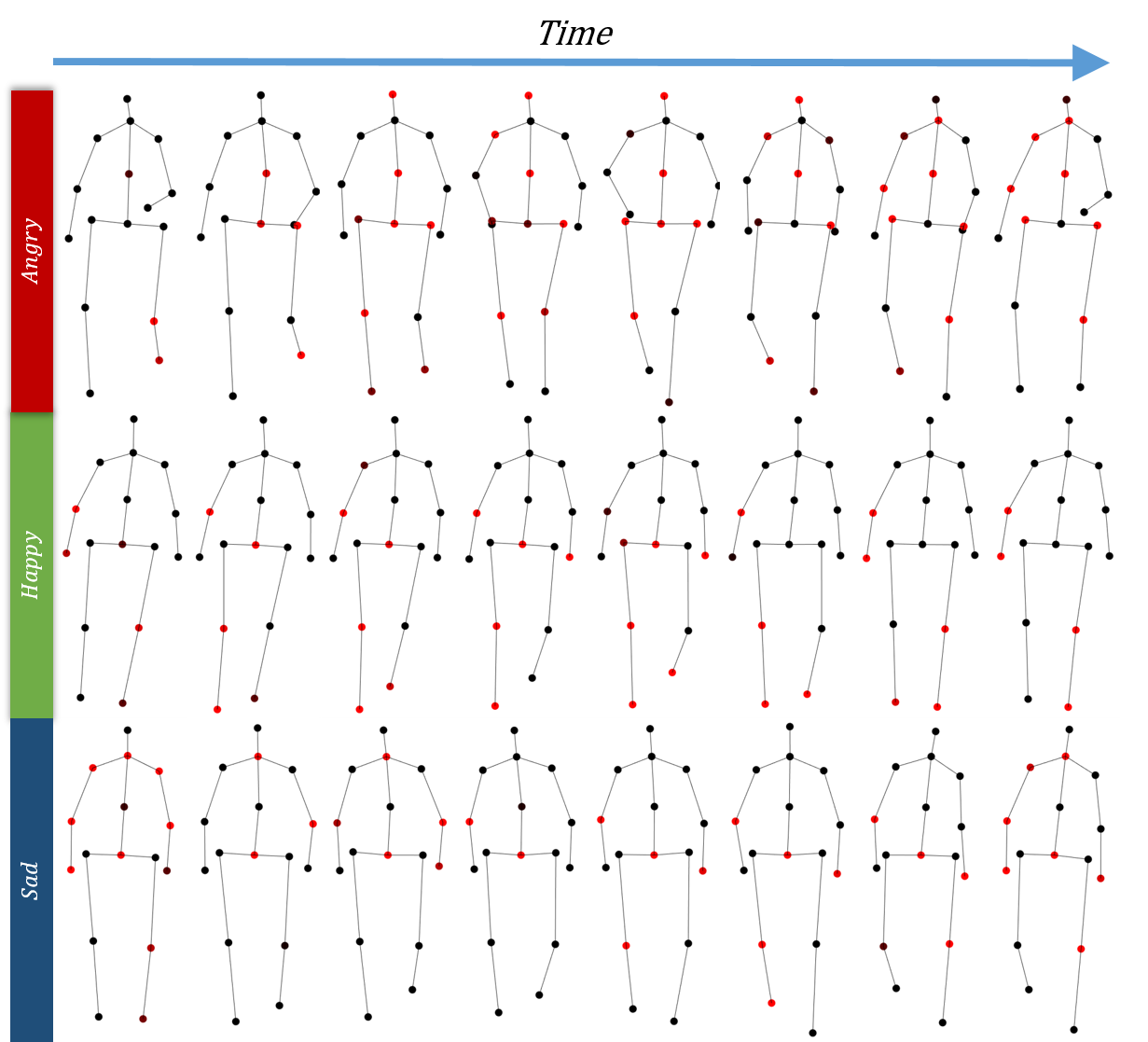}
    \caption{\textbf{\blue{Saliency Maps}}: \blue{We present the saliency maps for  one sample per emotion (angry, happy, and sad) as learned by the network for a single walk cycle.} The maps show activations on the joints during the walk cycle. Black represents no activation and red represents high activation. For all the emotion classes, the hand, feet and head joints have high activations, implying that the network deems these joints to be more important for determining the class. Moreover, the activation values on these joints for a high arousal emotion (\textit{e.g.}, angry) are higher than those for a low arousal emotion (\textit{e.g.}, sad), implying the network learns that higher arousal emotions lead to more vigorous joint movements.}
    \label{fig:saliency}
\end{figure*}

\blue{Additionally, we show the saliency maps given by the network in Figure~\ref{fig:saliency}. We selected one sample per emotion (angry, happy, and sad) and presented the postures in each row. For each gait (each row), we use eight sample frames corresponding to eight timesteps.  In each row, going from left to right, we show the evolution of the gait with time.  Each posture in the row shows the activation on the joints at the corresponding time step, as assigned by our network. Here, activation refers to the magnitude of the gradient of the loss w.r.t. an input joint (the joint's ``saliency") upon backpropagation through the learned network. Since all input data for our network are normalized to lie in the range $[0, 1]$, and the gradient of the loss function is smooth w.r.t. the inputs, the activation values for the saliency maps are within the $[0, 1]$ range. The joints are colored according to a gradient with $activation = 1$ denoting red and $activation = 0$ denoting black. The network uses these activation values of activated nodes in all the frames to determine the class label for the gait.}

\blue{We can observe from Figure~\ref{fig:saliency} that the network focuses mostly on the hand joints (observing arm swinging), the feet joints (observing stride), and the head and neck joints (observing head jerk). Based on the speed and frequency of the movements of these joints, the network decides the class labels. For example, the activation values on the joints for anger (Figure~\ref{fig:saliency} top row) are much higher than the ones for sadness (Figure~\ref{fig:saliency} bottom row), which matches with the psychological studies of how angry and sad gaits typically look. This shows that the features learned by the network are representative of the psychological features humans tend to use when perceiving emotions from gaits~\cite{karg2010recognition,michalak2009embodiment}. Additionally, as time advances, we can observe the pattern in which our network shifts attention to the different joints, i.e., the pattern in which it considers different joints to be more salient.  For example, if the right leg is moved ahead, the network assigns high activation on the joints in the right leg and left arm (and vice versa).}

\section{Emotional Walk \textit{(EWalk)} Dataset}\label{sec:dataset}
In this section, we describe our new dataset of videos of individuals walking. We also provide details about the perceived emotion annotations of the gaits obtained from this dataset.

\subsection{Data}
\begin{figure}[t]
    \centering
  \includegraphics[width=0.48\textwidth]{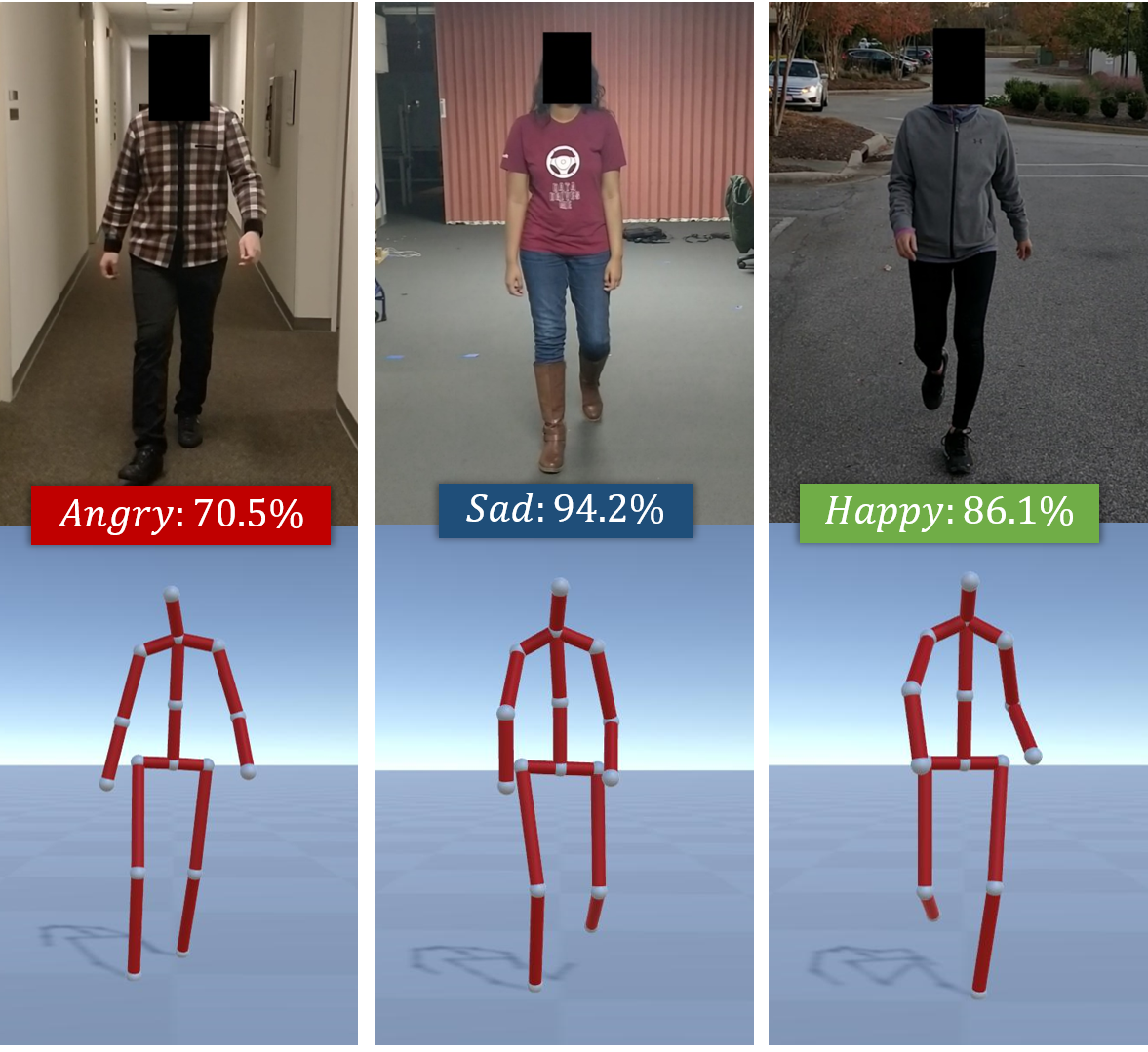}
  \caption{\textbf{EWalk Dataset:} We present the EWalk dataset containing RGB videos of pedestrians walking and the perceived emotion label for each pedestrian.}
  \label{fig:eWalk}
\end{figure}

The EWalk dataset contains $1384$ gaits with emotion labels from four basic emotions: happy, angry, sad, and neutral (Figure~\ref{fig:eWalk}). These gaits are either motion-captured or extracted from RGB videos. We also include synthetically generated gaits using state-of-the-art algorithms~\cite{xia2015realtime}. In addition to the emotion label for each gait, we also provide values of affective dimensions: valence and arousal.

\subsection{Video Collection}
We recruited $24$ subjects from a university campus. The subjects were from a variety of ethnic backgrounds and included $16$ male and $8$ female subjects. We recorded the videos in both indoor and outdoor environments. We requested that they walk multiple times with different walking styles. Previous studies show that non-actors and actors are both equally good at walking with different emotions~\cite{roether2009critical}. Therefore, to obtain different walking styles, we suggested that the subjects could assume that they are experiencing a certain emotion and walk accordingly. The subjects started  $7$m from a stationary camera and walked towards it. The videos were later cropped to include a single walk cycle.

\subsection{Data Generation}
Once we collect walking videos and annotate them with emotion labels, we can also use them to train generator networks to generate annotated synthetic videos. Generator networks have been applied for generating videos and joint-graph sequences of human actions such as walking, sitting, running, jumping, etc. Such networks are commonly based on either Generative Adversarial Networks (GANs)~\cite{gan} or Variational Autoencoders (VAEs)~\cite{vae}.

\begin{figure}[t]
    \centering
    \includegraphics[width=\columnwidth]{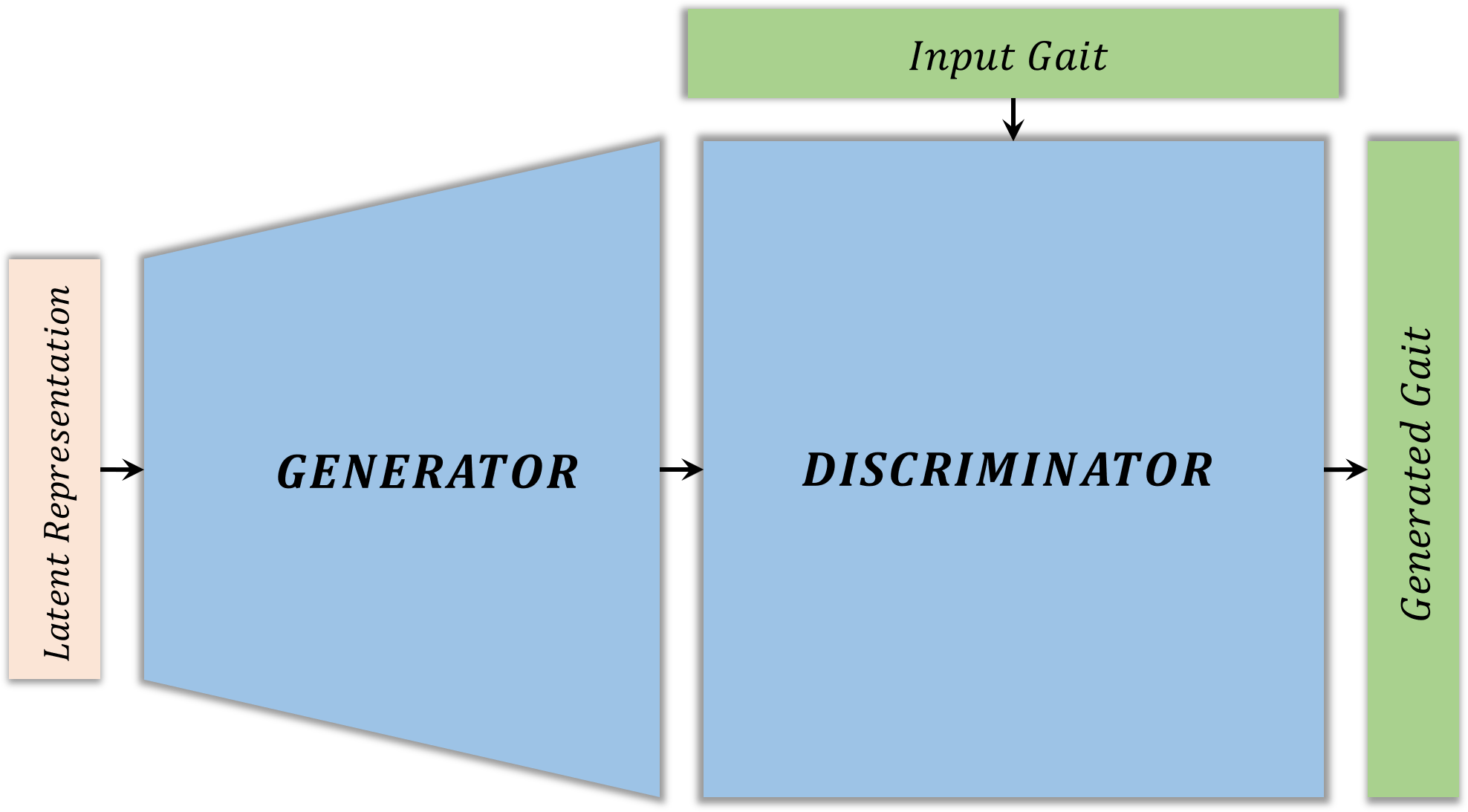}
    \caption{\textbf{Generative Adversarial Networks (GANs)}: The network consists of a generator that generates synthetic data from random samples drawn from a latent distribution space. This is followed by a discriminator that attempts to discriminate between the generated data and the real input data. The objective of the generator is to learn the latent distribution space of the real data whereas the objective of the discriminator is to learn to discriminate between the real data and the synthetic data generated by the generator. The network is said to be learned when the discriminator fails to distinguish between the real and the synthetic data.}
    \label{fig:gan}
\end{figure}

\begin{figure}[t]
    \centering
    \includegraphics[width=\columnwidth]{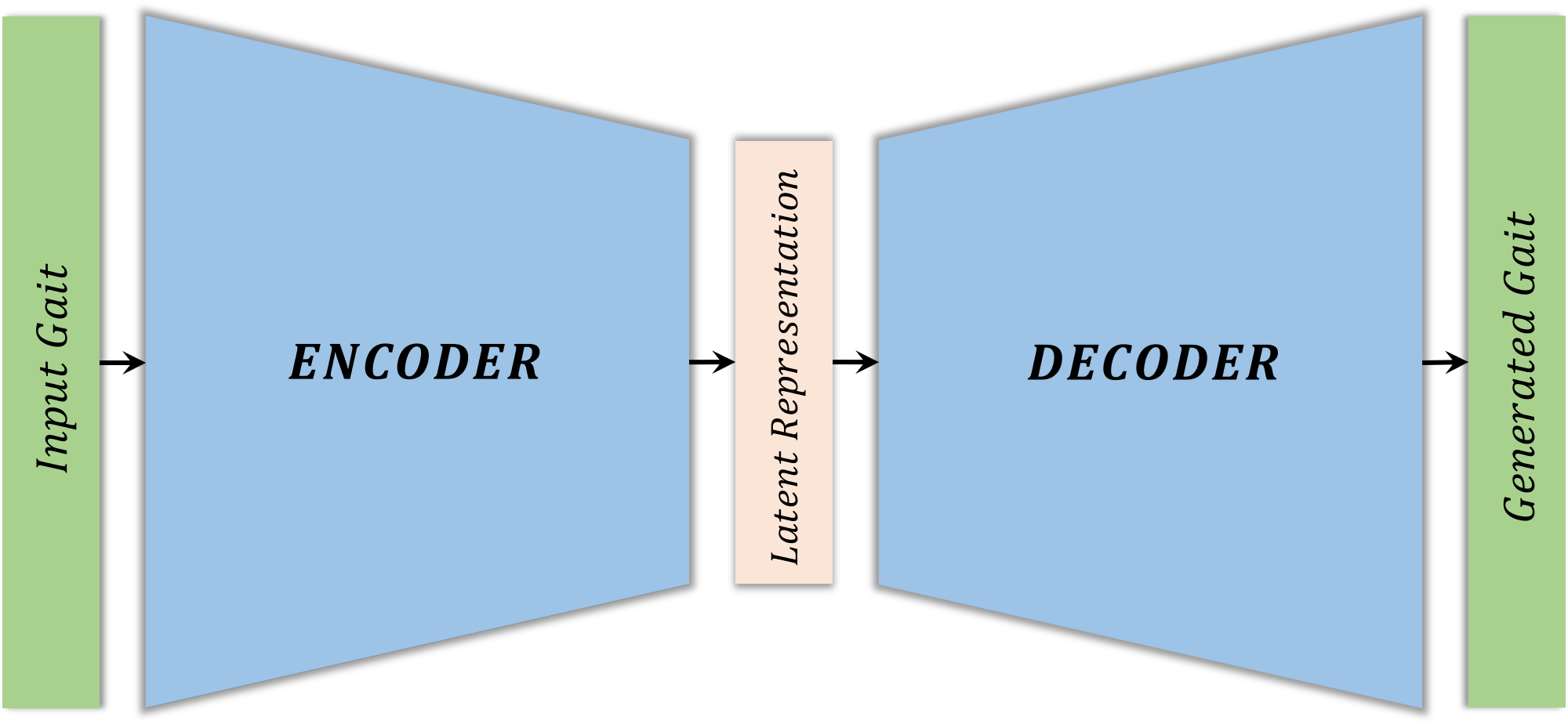}
    \caption{\textbf{Variational Autoencoders (VAEs)}: The encoder consists of an encoder that transforms the input data to a latent distribution space. This is followed by a discriminator that draws random samples from the latent distribution space to generate synthetic data. The objective of the overall network is then to learn the latent distribution space of the real data, so that the synthetic data generated by the decoder belongs to the same distribution space as the real data.}
    \label{fig:vae}
\end{figure}

GANs (Figure~\ref{fig:gan}) are comprised of a generator that generates data from random noise samples and a discriminator that discriminates between real data and the data generated by the generator. The generator is considered to be trained when the discriminator fails to discriminate between the real and the generated data.

VAEs (Figure~\ref{fig:vae}), on the other hand, are comprised of an encoder followed by a decoder. The encoder learns a latent embedding space that best represents the distribution of the real data. The decoder then draws random samples from the latent embedding space to generate synthetic data.

For temporal data such human action videos or joint-graph sequences, two different approaches are commonly taken. One approach is to individually generate each point in the temporal sequence (frames in a video or graphs in a graph sequence) respectively and then fuse them together in a separate network to generate the complete sequence. The methods in~\cite{twostep_method1, twostep_method2}, for example, use this approach. The network generating the individual points only considers the spatial constraints of the data, whereas the network fusing the points into the sequence only considers the temporal constraints of the data. The alternate approach is to train a single network by providing it both the spatial and temporal constraints of the data. For example, the approach used by Sijie et al.~\cite{stgcn}. The first approach is relatively more lightweight, but it does not explicitly consider spatial temporal inter-dependencies in the data, such as the differences in the arm swinging speeds between angry and sad gaits. While the latter approach does take these inter-dependencies into account, it is also harder to train because of these additional constraints.

\subsection{Analysis}

\begin{figure}[t]
    \centering
  \includegraphics[width=0.48\textwidth]{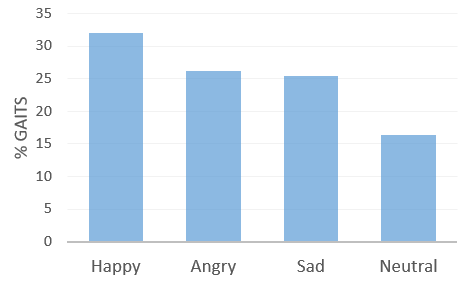}
  \caption{\textbf{Distribution of Emotion in the Datasets:} We present the percentage of gaits that are perceived as belonging to each of the emotion categories (happy, angry, sad, or neutral). We observe that our data is widely distributed.}
  \label{fig:dataDistribution}
\end{figure}

We presented the recorded videos to MTurk participants and obtained perceived emotion labels for each video using the method described in Section~\ref{sec:labeling}. Our data is widely distributed across the four categories with the \textit{Happy} category containing \blue{the largest number of gaits} ($32.07\%$) and the \textit{Neutral} category containing the smallest number of gaits with $16.35\%$ (Figure~\ref{fig:dataDistribution}).

\subsubsection{Affective Dimensions}\label{sec:affect}
We performed an analysis of the affective dimensions (i.e. valence and arousal). For this purpose, we used the participant responses to the questions about the happy, angry, and sad emotions. We did not use the responses to the question about the neutral emotion because it corresponds to the origin of the affective space and does not contribute to the valence and arousal dimensions. We performed a Principal Component Analysis (PCA) on the participant responses $[r^{happy}_i, r^{angry}, r^{sad}]$ and observed that the following two principal components describe $94.66\%$ variance in the data:
\begin{eqnarray}
 \begin{bmatrix} PC1 \\ PC2 \end{bmatrix} = \begin{bmatrix}
0.67 & -0.04 & -0.74 \\
-0.35 & 0.86 & -0.37 
\end{bmatrix}\label{eq:affectiveDimensions}
\end{eqnarray}

We observe that the first component with high values of the \textit{Happy} and \textit{Sad} coefficients represents the \textit{valence} dimension of the affective space. The second principal component with high values of the \textit{Anger} coefficient represents the \textit{arousal} dimension of the affective space. Surprisingly, this principal component also has a negative coefficient for the \textit{Happy} emotion. This is because a calm walk was often rated as happy by the participants, resulting in low arousal.

\subsubsection{Prediction of Affect}
We use the principal components from Equation~\ref{eq:affectiveDimensions} to predict the values of the \textit{arousal} and \textit{valence} dimensions. Suppose, the probabilities predicted by the Random Forest classifier are $p(h), p(a)$, and $p(s)$ corresponding to the emotion classes $happy$, $angry$, and $sad$, respectively. Then we can obtain the values of  $valence$ and $arousal$ as:
\begin{eqnarray}
valence = \begin{bmatrix}0.67 & -0.04 & -0.74 \end{bmatrix} \begin{bmatrix} p(h) & p(a) & p(s)\end{bmatrix} ^T \\
arousal = \begin{bmatrix}-0.35 & 0.86 & -0.37 \end{bmatrix} \begin{bmatrix} p(h) & p(a) & p(s)\end{bmatrix} ^T
\end{eqnarray}

\section{Conclusion, Limitations, and Future Work}
We presented a novel method for classifying perceived emotions of individuals based on their walking videos. Our method is based on learning deep features computed using LSTM and exploits psychological characterization to compute affective features. The mathematical characterization of computing gait features also has methodological implications for psychology research. This approach explores the basic psychological processes used by humans to perceive emotions of other individuals using multiple dynamic and naturalistic channels of stimuli. We concatenate the deep and affective features and classify the combined features using a Random Forest Classification algorithm. Our algorithm achieves an absolute accuracy of $80.07\%$,  which is an improvement of $24.60\%$ over vanilla LSTM (i.e., using only deep features) and offers an improvement of $13.85\%$ over state-of-the-art emotion identification algorithms. Our approach is the first approach to provide a real-time pipeline for emotion identification from walking videos by leveraging state-of-the-art 3D human pose estimation. We also present a dataset of videos (EWalk) of individuals walking with their perceived emotion labels. The dataset is collected with subjects from a variety of ethnic backgrounds in both indoor and outdoor environments. 

There are some limitations to our approach. The accuracy of our algorithm depends on the accuracy of the 3D human pose estimation and gait extraction algorithms. Therefore, emotion prediction may not be accurate if the estimated 3D human poses or gaits are noisy. Our affective computation requires joint positions from the whole body, but the whole body pose data may not be available if there are occlusions in the video. We assume that the walking motion is natural and does not involve any accessories (e.g., suitcase, mobile phone, etc.). As part of future work, we would like to collect more datasets and address these issues. We will also attempt to extend our methodology to consider more activities such as running, gesturing, etc. Finally, we would like to combine our method with other emotion identification algorithms that use human speech and facial expressions. \blue{We want to explore the effect of individual differences in the perception of emotions. We would also like to explore applications of our approach for simulating virtual agents with desired emotions using gaits.}




\bibliographystyle{IEEEtran}
\bibliography{template}

\newpage
\begin{IEEEbiography}[{\includegraphics[width=1in,height=1.25in,clip,keepaspectratio]{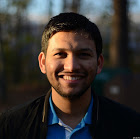}}]{Tanmay Randhavane}
Tanmay Randhavane is a graduate student at the Department of Computer Science, University of North Carolina, Chapel Hill, NC, 27514.
\end{IEEEbiography}

\vspace{-20pt}
\begin{IEEEbiography}[{\includegraphics[width=1in,height=1.25in,clip,keepaspectratio]{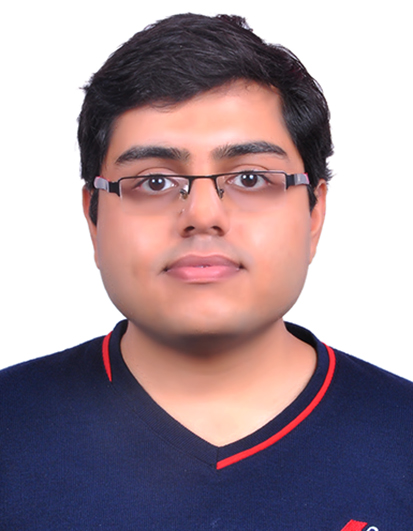}}]{Uttaran Bhattacharya}
Uttaran Bhattacharya is a graduate student at the Computer Science and Electrical \& Computer Engineering at the University of Maryland at College Park, MD, 20740.
\end{IEEEbiography}

\vspace{-20pt}
\begin{IEEEbiography}[{\includegraphics[width=1in,height=1.25in,clip,keepaspectratio]{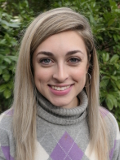}}]{Kyra Kapsaskis}
Kyra Kapsaskis is a full time research assistant in the Department of Psychology and Neuroscience, University of North Carolina, Chapel Hill, NC, 27514.
\end{IEEEbiography}

\vspace{-20pt}
\begin{IEEEbiography}[{\includegraphics[width=1in,height=1.25in,clip,keepaspectratio]{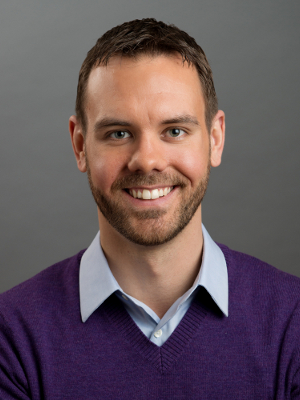}}]{Kurt Gray}
Kurt Gray is a Associate Professor with the Department of Psychology and Neuroscience, University of North Carolina, Chapel Hill, NC, 27514.
\end{IEEEbiography}

\vspace{-20pt}
\begin{IEEEbiography}[{\includegraphics[width=1in,height=1.25in,clip,keepaspectratio]{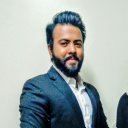}}]{Aniket Bera}
Aniket Bera is a Assistant Research Professor at the Department of Computer Science and University of Maryland Institute for Advanced Computer Studies, University of Maryland, College Park, 20740.
\end{IEEEbiography}

\vspace{-20pt}
\begin{IEEEbiography}[{\includegraphics[width=1in,height=1.25in,clip,keepaspectratio]{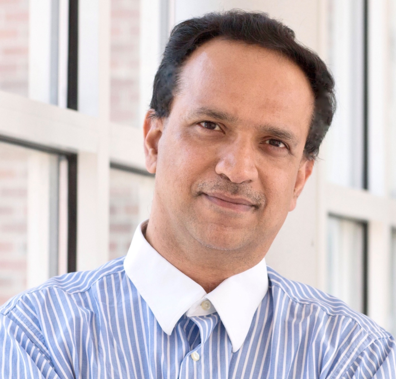}}]{Dinesh Manocha}
Dinesh Manocha is a Paul Chrisman Iribe Chair of Computer Science and Electrical \& Computer Engineering at the University of Maryland at College Park, MD, 20740.
\end{IEEEbiography}




\end{document}